\documentclass[runningheads]{llncs}

\usepackage{comment}
\usepackage{amsmath,amssymb}
\usepackage{color}
\usepackage{url}
\usepackage{hyperref}
\usepackage[per-mode=symbol]{siunitx}
\newif\ifreview
\reviewfalse

\ifreview
	\usepackage{lineno}

	\linenumbers
\fi

\usepackage{graphicx}
\usepackage{caption}
\usepackage{subcaption}

\newcommand{\picTwoAsym}[4]{
	\begin{figure*}[tb]
		\centering
		\begin{subfigure}[c]{0.64\textwidth}
			\centering		
			\includegraphics[width=0.9\textwidth]{#1-0}
			\subcaption{#2}	
			\label{fig:#1-0}	
		\end{subfigure}
		\begin{subfigure}[c]{0.35\textwidth}	
			\centering	
			\includegraphics[width=0.9\textwidth]{#1-1}
			\subcaption{#3}
			\label{fig:#1-1}	
		\end{subfigure}
		\caption{#4}
		\label{fig:#1}
	\end{figure*}
}

\usepackage{tabularx}
\usepackage{multirow}
\usepackage{booktabs}  
\usepackage{colortbl}  
\definecolor{LightGray}{rgb}{0.95,0.95,0.95}

\newcommand{\tblStatistics}{
\begin{table}[tb]
		\caption{
		Statistics for three different cohorts (all, CT or X-Ray data available). 
		For each cohort we give the statistics for the complete cohort and with respect to the hip fracture horizon with (w) and without (w/o) fracture (fx).
		$n$ represents the number of samples. Age is given as relative percentage for a specific range from $n$. 
		Training / Validation refer to the number of used images in the respective sets in the first fold.
		 - means that no data was used for training or validation.
		All other risk factors are given as mean (STD in brackets).
	}
	\label{tbl:statistics}
	\resizebox{\textwidth}{!}{
		\begin{tabular}{l c c c  c c c c c  c c c }
			& \multicolumn{3}{c}{Cohort All} & \multicolumn{5}{c}{Cohort CT} & \multicolumn{3}{c}{Cohort X-Ray}\\
			\cmidrule(r){2-4} \cmidrule(r){5-9} \cmidrule{10-12}
			
			& & \multicolumn{2}{c}{$t$ = 10 year} & & \multicolumn{2}{c}{$t$ = 10 year} & \multicolumn{2}{c}{$t$ = 5 year} & & \multicolumn{2}{c}{$t$ = 10 year} \\
			\cmidrule(r){3-4} \cmidrule(r){6-7} \cmidrule(r){8-9} \cmidrule{11-12}
			
			& all & w/o fx & w fx & all & w/o fx & w fx  & w/o fx & w fx & all & w/o fx & w fx \\
			
			\rowcolor{LightGray}
			
		    $n$ & 5994 & 4004 & 185 & 3165 & 2150 & 80 & 2198 & 32 & 3895 & 3108 & 89 \\
		    
		    Age [\%] \makecell{64-69\\70-74\\75-79\\80+} & \makecell{29.51\\28.50\\24.11\\17.88} & \makecell{36.69\\31.87\\21.98\\9.47} & \makecell{10.27\\16.22\\33.51\\40.00} 
 & \makecell{30.74\\28.44\\23.70\\17.12} & \makecell{37.86\\32.00\\21.44\\8.70} & \makecell{16.25\\15.00\\37.50\\31.25}  & \makecell{37.44\\31.71\\21.84\\9.01} & \makecell{12.50\\9.38\\34.38\\43.75}   & \makecell{32.99\\30.40\\23.90\\12.71} & \makecell{37.32\\31.66\\22.30\\8.72} & \makecell{15.73\\19.10\\39.33\\25.84}  \\ 
 
 \rowcolor{LightGray}
 
		      Height [m] & \makecell{1.74\\(0.07)}  & \makecell{1.75\\(0.07)}   & \makecell{1.73\\(0.06)}  & \makecell{1.74\\(0.07)}  & \makecell{1.75\\(0.07)}   & \makecell{1.74\\(0.06)}  & \makecell{1.75\\(0.07)}   & \makecell{1.72\\(0.06)}  & \makecell{ 1.74\\(0.07)}  & \makecell{1.74\\(0.07)}   & \makecell{1.74\\(0.06)}    \\

BMI [$\frac{kg}{m^2}$] & \makecell{25.90\\(3.65)}  & \makecell{26.00\\(3.54)}   & \makecell{25.11\\(3.63)}   & \makecell{25.80\\(3.60)}  & \makecell{25.89\\(3.48)}   & \makecell{24.77\\(3.38)}   & \makecell{ 25.86\\(3.49)}   & \makecell{24.68\\(3.02)}   & \makecell{25.86\\(3.57)}  & \makecell{25.90\\(3.48)}   & \makecell{24.86\\(3.65)}   \\

\midrule

\rowcolor{LightGray}

\makecell{Femoral aBMD\\$[\frac{g}{cm^2}]$}  & \makecell{0.78\\(0.13)}  & \makecell{0.79\\(0.13)}   & \makecell{0.66\\(0.11)}    & \makecell{0.78\\(0.13)}  & \makecell{0.79\\(0.13)}   & \makecell{0.65\\(0.08)}   & \makecell{0.79\\(0.13)}   & \makecell{0.62\\(0.08)}   & \makecell{ 0.79\\(0.13)}  & \makecell{0.79\\(0.12)}   & \makecell{0.68\\(0.10)}     \\

\makecell{Spine aBMD\\ $[\frac{g}{cm^2}]$} & \makecell{1.07\\(0.19)}  & \makecell{1.07\\(0.18)}   & \makecell{1.01\\(0.19)}   & \makecell{1.07\\(0.19)}  & \makecell{1.06\\(0.18)}   & \makecell{1.00\\(0.16)}   & \makecell{1.06\\(0.18)}   & \makecell{0.98\\(0.19)}  & \makecell{1.07\\(0.18)}  & \makecell{1.07\\(0.18)}   & \makecell{1.00\\(0.20)}   \\

\rowcolor{LightGray}

Avg. TBS & \makecell{1.23\\(0.13)}  & \makecell{1.23\\(0.13)}   & \makecell{1.19\\(0.13)}   & \makecell{ 1.23\\(0.13)}  & \makecell{1.24\\(0.12)}   & \makecell{1.20\\(0.12)}   & \makecell{1.24\\(0.12)}   & \makecell{1.17\\(0.13)}     & \makecell{1.24\\(0.12)}  & \makecell{1.24\\(0.12)}   & \makecell{1.19\\(0.12)}     \\

\frax [\%] & \makecell{4.14\\(4.39)}  & \makecell{3.14\\(3.07)}   & \makecell{7.04\\(5.67)}   & \makecell{4.01\\(4.28)}  & \makecell{3.04\\(2.93)}   & \makecell{6.29\\(5.85)}   & \makecell{3.07\\(2.94)}   & \makecell{8.53\\(7.97)}   & \makecell{3.54\\(3.56)}  & \makecell{3.07\\(2.89)}   & \makecell{5.76\\(4.87)}    \\

\rowcolor{LightGray}

\makecell{\frax (w. aBMD)\\ $ [\%] $ } & \makecell{ 4.45\\(5.54)}  & \makecell{3.38\\(4.04)}   & \makecell{10.86\\(9.09)}  & \makecell{4.33\\(5.41)}  & \makecell{3.35\\(3.95)}   & \makecell{10.47\\(8.92)}   & \makecell{  3.45\\(4.12)}   & \makecell{14.38\\(9.41)}  & \makecell{3.85\\(4.74)}  & \makecell{3.34\\(3.95)}   & \makecell{9.41\\(8.56)}     \\
		    
		    \midrule

	        Training & - & - & - & - & 3403 & 128 & 3478 & 53 & - & 4353 & 107 \\
	       \rowcolor{LightGray}
	        
	        Validation & - & - & - & - & 790 & 27 & 810 & 7 & - &  1086 & 35 \\
			
		\end{tabular}
	}

\end{table}
}

\newcommand{\tblHorExtended}{
\begin{table}[tb]
		\caption{
		Extended results on CTN data and with 5 year horizon
	}
	\label{tbl:results}
	\resizebox{\textwidth}{!}{
		\begin{tabular}{l c c c c c  c  c c c }
			& & \multicolumn{5}{c}{Opportunistic Setting} & \multicolumn{3}{c}{Non-Opportunistic Setting}\\
			\cmidrule(r){3-7} \cmidrule(r){8-10}
			 &  & Base & Multiple & \makecell{GAP\\(CTN / CT / X-Ray)}  & \makecell{GAP + Base\\(CTN / CT / X-Ray)} & \makecell{GAP + Multiple\\(CTN / CT / X-Ray)} & \frax$^\dagger$ &  \makecell{aBMD +\\Base} & \makecell{\frax +\\Base} & \makecell{TBS +\\Base} & \frax + aBMD$^\dagger$\\
			
		    & \\
			
			&   & \makecell{60.89 $\pm$\\0.73} & \makecell{61.86 $\pm$\\0.88} & \makecell{80.87 $\pm$ 0.13/\\82.58 $\pm$ 0.21/\\77.24 $\pm$ 0.30} & \makecell{81.04 $\pm$ 0.17/\\ 82.48 $\pm$ 0.24/\\77.81 $\pm$ 0.38}  & \makecell{81.66 $\pm$ 0.23/\\82.52 $\pm$ 0.28/\\78.41 $\pm$ 0.33} & \makecell{68.49 $\pm$\\0.71} & \makecell{77.28 $\pm$\\0.58} & \makecell{60.50 $\pm$\\0.90} \\ 
			
				\midrule
			
			\makecell{CTN +\\Multiple}  & \makecell{81.66\\$\pm$ 0.23} &  \textbf{+20.77}&  \textbf{+19.80}&  \textbf{+0.78}/\textbf{-0.92}/\textbf{+4.41}  &  \textit{+0.61}/\textit{-0.82}/\textbf{+3.85}   &  - /\textit{-0.86}/\textbf{+3.24}  &  \textbf{+13.16}&  \textbf{+4.38} &  \textbf{+21.16}\\ 
            \makecell{CT +\\Multiple} & \makecell{82.52\\$\pm$ 0.28} &  \textbf{+21.63} &  \textbf{+20.66}&  \textbf{+1.64}/-0.06/\textbf{+5.28} &  \textbf{+1.47}/ +0.04/\textbf{+4.71}  &  \textit{+0.86}/ - /\textbf{+4.10}   &   \textbf{+14.03}&  \textbf{+5.24} &  \textbf{+22.02}\\ 
            \makecell{X-Ray +\\Multiple} & \makecell{78.41\\$\pm$ 0.33} & \textbf{+17.53} & \textbf{+16.56} & \textbf{-2.46}/\textbf{-4.17}/ \textit{+1.17} & \textbf{-2.63}/\textbf{-4.07}/ +0.61 & \textbf{-3.24}/\textbf{-4.10}/ - & \textbf{+9.92} & +1.14 & \textbf{+17.91}

		\end{tabular}
	}
	\subcaption{10 year hip fracture risk}
	
	\resizebox{\textwidth}{!}{
		\begin{tabular}{l c  c c c c c  c c c }
			 & & \multicolumn{5}{c}{Opportunistic Setting} & \multicolumn{3}{c}{Non-Opportunistic Setting}\\
			\cmidrule(r){3-7} \cmidrule(r){8-10}
			 &  & Base & Full & \makecell{GAP\\(CTN / CT)}  & \makecell{GAP + Base\\(CTN / CT)} & \makecell{GAP + Full\\(CTN / CT)} & \makecell{aBMD +\\Multiple} & \makecell{\frax +\\Multiple} & \makecell{TBS +\\Multiple}\\
			
		    & \\
			
			&   & 64.88 $\pm$ 0.80 & 63.46 $\pm$ 1.12 & \makecell{88.49 $\pm$ 0.32/\\95.65 $\pm$ 0.27} & \makecell{89.35 $\pm$ 0.35/\\95.65 $\pm$ 0.37} & \makecell{89.62 $\pm$ 0.36/\\95.55 $\pm$ 0.30} &  71.63 $\pm$ 0.77 & 79.86 $\pm$ 0.86 & 61.76 $\pm$ 0.94 &\\ 
			
			\midrule

 \makecell{CTN +\\Multiple}   & \makecell{89.62\\$\pm$ 0.36} &  \textbf{+24.73}&  \textbf{+26.16}&  +1.13/\textbf{-6.03}  &  +0.26/\textbf{-6.04}   &  - /\textbf{-5.93}  &  \textbf{+17.99}&  \textbf{+9.76} &  \textbf{+27.86}\\ 
 \makecell{CT +\\Multiple}   & \makecell{95.55\\$\pm$ 0.30} &  \textbf{+30.66}&  \textbf{+32.09}&  \textbf{+7.06}/-0.10 &  \textbf{+6.20}/-0.11  &  \textbf{+5.93}/ - &   \textbf{+23.92}&  \textbf{+15.69}& \textbf{ +33.79}\\

		\end{tabular}
	}
	
	\subcaption{5 year hip fracture risk}
}

\newcommand{\tblHor}{
\begin{table}[tb]
		\caption{
		Comparison fracture risk prediction --
		columns: different inputs which were used to train FORM;
		rows: different cohorts.
		All scores are given as mean val AUC $\pm$ SE. 
		Significant differences between the first and the other columns are marked italic (p $<$ 0.05) or bold  (p $<$ 0.01).
		$\dagger$ input not used for FORM
	}
	\label{tbl:results}
	\resizebox{\textwidth}{!}{
		\begin{tabular}{l c | c c c c | c c  }
			 Cohort & GAP + Multiple  &  Base & Multiple & GAP  & GAP + Base & \frax$\dagger$ \\

				\midrule
			
\makecell{X-Ray } & 78.41 $\pm$ 0.33  & \textbf{66.38 $\pm$ 1.76} & \textbf{69.67 $\pm$ 0.99} & \textit{77.24 $\pm$ 0.30} & 77.81 $\pm$ 0.38 &  77.43\\
            \makecell{CT } & 82.67 $\pm$ 0.21 &  \textbf{60.89 $\pm$ 0.73} &  \textbf{67.03 $\pm$ 0.93}&  82.58 $\pm$ 0.21 &  82.48 $\pm$ 0.24  &   75.94 \\

		\end{tabular}
	}
	
	\subcaption{Non-densitometric Settings}
	
		\resizebox{\textwidth}{!}{
		\begin{tabular}{l c |c c c | c c c c c c c}
			 Cohort & GAP + Multiple & \makecell{aBMD + Base} & \makecell{\frax + aBMD + Base} & \makecell{TBS + Base} & \frax + aBMD$\dagger$\\

				\midrule
			
 \makecell{X-Ray} & 78.41 $\pm$ 0.33  & \textbf{76.55 $\pm$ 0.89} & \textbf{81.50 $\pm$ 0.83} & \textbf{72.66 $\pm$ 1.34} & 80.92\\
            \makecell{CT} & 82.67 $\pm$ 0.21 &   \textbf{71.82 $\pm$ 0.50} &  \textbf{81.08 $\pm$ 0.34} &  \textbf{71.56 $\pm$ 0.39} &  79.19\\

		\end{tabular}
	}
	
	\subcaption{Densitometric Settings}

\end{table}
}

\newcommand{\tblCross}{
\begin{table}[tb]
		\caption{
		Cross-validation results (mean val. AUC $\pm$ STD) --
		columns: different methods / inputs;
		rows: cohort. Bold: results within a one percent margin of the best for each cohort.
	}
	\label{tbl:cross}
	\resizebox{\textwidth}{!}{
		\begin{tabular}{l c  c c c c c c c  }
	 &  \multicolumn{3}{c}{FORM} & \multicolumn{3}{c}{Cox} &  \frax  \\ 
	 \cmidrule(r){2-4} \cmidrule(r){5-7}
			 Cohort & GAP  & GAP + Base &  GAP + Multiple & GAP  & GAP + Base &  GAP + Multiple  & \\

				\midrule
			
			\makecell{X-Ray } & \textbf{81.57 $\pm$ 3.13} & \textbf{81.09 $\pm$ 3.18} & \textbf{81.44 $\pm$ 3.11} & 61.14 $\pm$ 16.80 & 70.26 $\pm$ 5.71 & 70.19 $\pm$ 6.58  &  74.72 $\pm$ 7.21 \\

            \makecell{CT } & 77.53 $\pm$ 5.81 &  \textbf{80.66 $\pm$ 3.75}  &  \textbf{81.04 $\pm$ 5.54} & 67.56 $\pm$ 23.97 & 73.69 $\pm$ 9.22 & 75.35 $\pm$ 9.11 &   74.74 $\pm$ 5.70\\

			\midrule

		\end{tabular}
	}

\end{table}
}

\newcommand{\tblRiskFactors}{
\begin{table}[tbh]
		\caption{
	 Overview about individual elements of the risk factor groups
	}
	\label{tbl:risk}
	\resizebox{\textwidth}{!}{
		\begin{tabular}{l l }
		
		Risk factor group & Individual Elements \\
		
		\midrule
		Base & age, BMI, height, fat mass, clinic \\
		Multiple  &  \makecell[l]{Base +  previous fractures spine/hip, osteoporosis in parents,  fall history, professional background,\\
		previous diseases like cancer or hypertension  and smoking and alcohol habits} \\
		aBMD  & spinal and femoral aBMD \\
		\frax & \frax 10 year risk  for hip and major osteoporotic fracture with and without usage of aBMD \\
		TBS & Average TBS across spine \\
		
		\end{tabular}
	}
\end{table}
}

\newcommand{\tblAblation}{
\begin{table}[thb]
		\caption{
	 Ablation of hyperparameters on CTN data for 10 and 5 year hip fracture risk --
	 Significant differences to the baseline are marked italic  (p $<$ 0.05) or bold  (p $<$ 0.01).
	}
	\resizebox{\textwidth}{!}{
		\begin{tabular}{l c  c c c c c  c c c }
			& \multicolumn{3}{c}{10 year risk} & \multicolumn{3}{c}{5 year risk}\\
			\cmidrule(r){2-4} \cmidrule(r){5-7}
			
			&  GAP & GAP + Base & GAP + Multiple  &  GAP & GAP + Base & GAP + Multiple \\
			 
			CTN & 80.87 $\pm$ 0.13 & 81.04 $\pm$ 0.17 & 81.66 $\pm$ 0.23 & 88.49 $\pm$ 0.32 & 89.35 $\pm$ 0.35 & 89.62 $\pm$ 0.36 \\ 
			\midrule
 Dropout rate 0.4 &   \textit{-0.34} (80.53 $\pm$ 0.18)&  -0.18 (80.87 $\pm$ 0.27)&  \textit{-0.98} (80.68 $\pm$ 0.33)	&   -0.47 (88.01 $\pm$ 0.37)&  -1.14 (88.21 $\pm$ 0.43)&  \textit{-2.00} (87.61 $\pm$ 0.44)\\ 
 Min-Max-Normalization   &   +0.05 (80.92 $\pm$ 0.19)&  -0.04 (81.00 $\pm$ 0.17)&  \textit{-0.65} (81.00 $\pm$ 0.25) &   -0.15 (88.34 $\pm$ 0.43)&  -0.99 (88.37 $\pm$ 0.37)&  -1.21 (88.41 $\pm$ 0.54)\\ 
 Smaller MLP &   -0.40 (80.47 $\pm$ 0.40)&  -0.13 (80.92 $\pm$ 0.24)&  \textit{-0.70} (80.95 $\pm$ 0.25) & +0.05 (88.54 $\pm$ 0.53)&  -0.24 (89.12 $\pm$ 0.52)&  -1.91 (87.70 $\pm$ 0.58)\\ 
 Larger MLP &   -0.02 (80.86 $\pm$ 0.14)&  \textbf{+0.57} (81.61 $\pm$ 0.21)&  -0.20 (81.46 $\pm$ 0.31)&   +0.42 (88.91 $\pm$ 0.28)&  \textbf{-1.83} (87.52 $\pm$ 0.25)&  \textbf{-2.32} (87.30 $\pm$ 0.32)\\  
No input merging  &   +0.01 (80.88 $\pm$ 0.22)&  -0.19 (80.86 $\pm$ 0.19)&  \textbf{-0.78} (80.87 $\pm$ 0.19)&   +0.72 (89.21 $\pm$ 0.41)&  \textit{-1.60} (87.75 $\pm$ 0.31)&  \textbf{-3.04} (86.58 $\pm$ 0.28) \\
Input merging with s = 1  &   -0.40 (80.47 $\pm$ 0.28)&  +0.10 (81.15 $\pm$ 0.25)&  -0.32 (81.34 $\pm$ 0.29)&   +0.12 (88.61 $\pm$ 0.39)&  \textit{-1.83} (87.52 $\pm$ 0.45)&  -1.90 (87.72 $\pm$ 0.51)  
			
		\end{tabular}
	}
	\label{tbl:ablation}
\end{table}
}

\newcommand{\tblDatasets}{
\begin{table}[thb]
		\caption{
	 Overview of reason for exclusion of patients -- We start for all cohorts with the total study population. The reasons for certain exclusions and number of excluded cases are given in white rows and the left over people are given in the grey rows.
	}
	\resizebox{\textwidth}{!}{
	\begin{tabular}{c c c c  c c c c c  c c c }
			& \multicolumn{3}{c}{Cohort All} & \multicolumn{5}{c}{Cohort CT} & \multicolumn{3}{c}{Cohort X-Ray}\\
			\cmidrule(r){2-4} \cmidrule(r){5-9} \cmidrule{10-12}
			
			& & \multicolumn{2}{c}{$t$ = 10 year} & & \multicolumn{2}{c}{$t$ = 10 year} & \multicolumn{2}{c}{$t$ = 5 year} & & \multicolumn{2}{c}{$t$ = 10 year} \\
			\cmidrule(r){3-4} \cmidrule(r){6-7} \cmidrule(r){8-9} \cmidrule{11-12}
			
			& all & w/o fx & w fx & all & w/o fx & w fx  & w/o fx & w fx & all & w/o fx & w fx \\
			
			\rowcolor{LightGray}
			
			Start & 5994 & 5994 & 5994 & 5994 & 5994 & 5994 & 5994 & 5994 & 5994 & 5994 & 5994 \\
			
			\makecell{Not all risk\\factors present} & N/A & N/A & N/A & 52 & 52 & 52 & 52 & 52 & 52 & 52 & 52  \\
			\rowcolor{LightGray}
			& 5994 & 5994 & 5994 & 5942 & 5942 & 5942 & 5942 & 5942 & 5942 & 5942 & 5942   \\
			
			\makecell{No Image\\Data} & N/A & N/A & N/A & 2670 & 2670 & 2670 & 2670 &  2670 & 1759 & 1759 & 1759\\
			\rowcolor{LightGray}
			& 5994 & 5994 & 5994 & 3272 & 3272 & 3272 & 3272 & 3272 & 4183 & 4183 & 4183  \\
			
			\makecell{Censored} & N/A & 1805 & 1805 & N/A & 970 & 970 & 970 & 970 & N/A & 762 & 762 \\
			\rowcolor{LightGray}
			& 5994 & 4189 & 4189 & 3272 & 2302 & 2302 & 2302 & 2302 & 4183 & 3421 & 3421\\
			
			\makecell{Both Hips\\not usable} & N/A  & N/A & N/A & 107 & 72 & 72 & 72 & 72 & 288 & 224 & 224   \\
			\rowcolor{LightGray}
			& 5994 & 4189 & 4189 & 3165 & 2230 & 2230 &  2230 & 2230 & 3895 & 3197 & 3197 \\
			
			Wrong Class & N/A & 185 & 4004 & N/A & 80 & 2150 & 32 & 2198 & N/A & 89 & 3108  \\
			\rowcolor{LightGray}
		    $n$ & 5994 & 4004 & 185 & 3165 & 2150 & 80 & 2198 & 32 & 3895 & 3108 & 89 \\

		\end{tabular}
	}
	\label{tbl:datasets}
\end{table}
}

\usepackage{makecell}

\usepackage{placeins}

\def\frax{FRAX\textsuperscript{\raisebox{3pt}{\scalebox{.4}{\textregistered}\:}}}

\begin{document}

\def\SubNumber{xx}

\title{Opportunistic hip fracture risk prediction in Men from X-ray: Findings from the Osteoporosis in Men (MrOS) Study}

\ifreview
	\institute{Paper ID \SubNumber}
\else
	\titlerunning{Opportunistic hip fracture riks prediction}

	\author{Lars Schmarje\inst{1,*} \and
	Stefan Reinhold\inst{1} \and
	Timo Damm\inst{2} \and
	Eric Orwoll\inst{3} \and
	Claus-C. Glüer\inst{2} \and
	Reinhard Koch\inst{1}}
	
	\authorrunning{L. Schmarje et al.}
	
	\institute{MIP, Computer Science, Kiel University 
	\email{\{las,sre,rk\}@informatik.uni-kiel.de} \and
    MOINCC, Kiel University, Germany \email{\{timo.damm,glueer\}@rad.uni-kiel.de} \and
	Oregon Health \& Science University, United States 
	\email{orwoll@ohsu.edu}\\
	$^*$ Corresponding author: \email{las@informatik.uni-kiel.de}}
\fi

\maketitle              
\begin{abstract}
Osteoporosis is a common disease that increases fracture risk. Hip fractures, especially in elderly people, lead to increased morbidity, decreased quality of life and increased mortality. Being a silent disease before fracture, osteoporosis often remains undiagnosed and untreated.
Areal bone mineral density (aBMD) assessed by dual-energy X-ray absorptiometry (DXA) is the gold-standard method for osteoporosis diagnosis and hence also for future fracture prediction (prognostic).
However, the required special equipment is not broadly available everywhere, in particular not to patients in developing countries.
We propose a deep learning classification model (FORM) that can directly predict hip fracture risk from either plain radiographs (X-ray) or 2D projection images of computed tomography (CT) data.
Our method is fully automated and therefore well suited for opportunistic screening settings, identifying high risk patients in a broader population without additional screening.
FORM was trained and evaluated on X-rays and CT projections from the Osteoporosis in Men (MrOS) study.
3108 X-rays (89 incident hip fractures) or 2150 CTs (80 incident hip fractures) with a 80/20 split (training / validation) were used.
We show that FORM can correctly predict the 10-year hip fracture risk with a validation AUC of 81.44\% $\pm$ 3.11\% / 81.04\% $\pm$ 5.54\% (mean $\pm$ STD) including additional information like age, BMI, fall history and health background across a 5-fold cross validation on the X-ray and CT cohort, respectively.
Our approach significantly (p $<$ 0.01) outperforms previous methods like Cox Proportional-Hazards Model and \frax  with 70.19 $\pm$ 6.58 and 74.72 $\pm$ 7.21 respectively on the X-ray cohort.
Our model outperform on both cohorts hip aBMD based predictions (validation AUC 82.67\% $\pm$ 0.21\% vs. 71.82\% $\pm$ 0.50\%   and 78.41\% $\pm$ 0.33 vs. 76.55\% $\pm$ 0.89\%).
We are confident that FORM can contribute on improving osteoporosis diagnosis at an early stage.

\end{abstract}
\section{Introduction}

Osteoporosis is a wide-spread systemic disease that leads to deterioration of bone mass and micro structure and subsequently to decreased bone strength inducing an increased fracture risk \cite{salari2021global}.
According to the United States Preventive Services Task Force, the lifetime risk of an osteoporotic fracture is about 50\% in women and about 20\% - 25\% in men \cite{us2011screening,prasad2021chronic}.
While osteoporosis affects all bones, fractures of the spine and hip are the most frequent. 
Especially hip fractures lead to increased morbidity, decreased quality of life and increased mortality --- 20\% of osteoporotic hip fractures lead to death within six month\cite{ebeling2014osteoporosis}.
Being a silent disease before fracture, osteoporosis often remains undiagnosed and consequently untreated.
Especially in men, only about 2\% are diagnosed before fracture \cite{prasad2021chronic}.

The gold-standard method for osteoporosis diagnosis is based on areal bone mineral density (aBMD) assessed by dual-energy X-ray absorptiometry (DXA). This modality is in general broadly available to patients in many countries world-wide - with some degree of uneven distribution among industrial nations. In developing countries in African and South America and the Middle East, the availability is poor \cite{johnell2006estimate,hamidi}.
More elaborate methods like volumetric bone mineral density (vBMD) assessed by quantitative computed tomography (QCT) or finite element modeling (FEM) of bone strength, either based on QCT or DXA, have shown to be superior to standard aBMD \cite{Yang2018fem,langsetmo2018volumetric,schousboe2016incident,wang2012prediction,black2008proximal}.
However, all these method either require special equipment, protocols or domain experts and 
the prognosis of osteoporotic fractures is an even more challenging and labor-intensive task.

In this paper, we focus on fracture prognosis in an opportunistic screening scenario:
whenever radiographic imaging is available an automated method inspects the image for indicators of  possible future fractures.
Patients with high fracture risk could be advised to see a specialist to confirm the risk and possibly initiate preventive actions.
Due to their outstanding capacity to learn task-relevant image features such methods - in particular convolutional neural networks (CNN) – have outperformed “classical” machine learning algorithms in many image analysis tasks\cite{fixmatch,foc,santarossa2022medregnet,dc3,grossmann2022beyond,schmarje2022benchmark}.
We predict the risk of future fractures (prognostics), in contrast to detecting acute osteoporosis or incident fractures (diagnostics).

The goal is to develop a pipeline that can be used for opportunistic screening and hence beneficially leverage additional risk factors.
For this purpose we propose a two-stage deep learning based classification method that is able to predict the 10-year fracture risk using only X-ray or CT scans and optionally case history as inputs.
We train and evaluate our method on a dataset from the Osteoporotic Fractures in Men (MrOS\footnote{The Osteoporotic Fractures in Men (MrOS) Study: \url{https://mrosonline.ucsf.edu}}) study.
We restrict our main evaluation to information (e.g. age, weight, height, etc.) that would be collectible in this setting;
other information (e.g. aBMD) is only included for comparison.

\textbf{Our key contributions are}:
(1) a fully automated system that can be used in an prognostic opportunistic screening scenario,
(2) val AUC results of 81.44\% and 81.04\% on the X-ray and CT cohort respectively.
(3) we significantly outperform previous methods like Cox Proportional-Hazards Model and \frax in the opportunistic use case and achieve improved or competitive results for non-opportunistic settings,
(4) beneficial integration of clinical risk factors with image based features into a deep learning pipeline.

\picTwoAsym{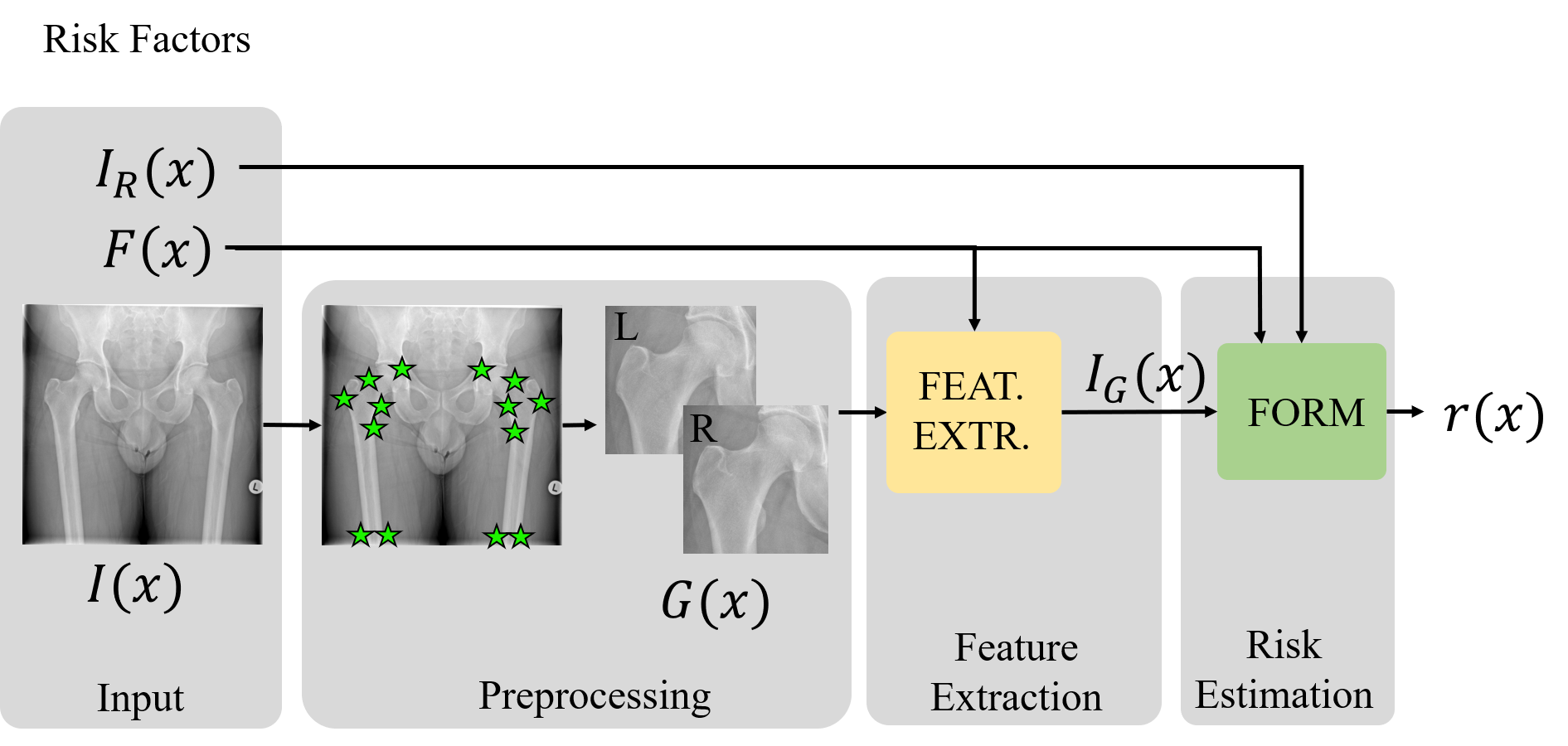}{Overview Pipeline}{Overview of the used models}{Illustration\protect\footnotemark of proposed pipeline.
(a) Inputs and pipeline stages: preprocessing, feature extraction and risk estimation. (b) Detailed view of models in (a) (yellow, green).
Parameter of fully connected (FC) layers: number of hidden neurons.
Further information can be found in \autoref{sec:method}.}

\subsection{Related Work}

In the past numerous risk factors for osteoporotic fractures were identified.\footnotetext{Example image and key points only for illustrative purpose; image source \url{https://radiopaedia.org/cases/normal-hip-x-rays}}
Among them are increased age, low body mass index (BMI), previous fragility fractures, smoking or alcohol intake.
While aBMD alone has shown to be a not sufficiently sensitive predictor for screening applications\cite{kannis2008frax}, the combination with other risk factors (RF) is more promising.
In \cite{schousboe2014vertrebralfractures} Schousboe et al.  found that additional RF are better than a model using only aBMD and age for vertebrae fracture prediction.
Elaborate statistical shape and density modeling based on volumetric QCT data has proven to be superior to DXA based aBMD models\cite{bredbenner2014fractureRisk} for hip fracture prognosis.
Hippisley-Cox et al. proposed the QFractureScores algorithm \cite{hippisley2009cox} to predict the 10-year fracture risk. 
\frax \cite{kannis2008frax} is a fracture risk assessment tool that uses various RF with or without additional aBMD measurements to predict a 10-year fracture risk.
The National Osteoporosis Foundation included \frax in its guidelines to recommend aBMD measurements or even treatment based on predicted fracture risk\cite{watts2011fracture}.
Su et al. \cite{Su2019cart} used classification and regression trees (CART) on common RF to predict fracture risk on the MrOS data and found a slight improvement over \frax based predictions.
Treece et al. \cite{treece2015predicting} used cortical bone mapping (CBM) to predict osteoporotic fractures in the MrOS study. They found that adding CBM to aBMD can improve fracture prognosis performance.
Most of these methods do, however, require special protocols, modalities or in-depth interviews of the patient that might not be applicable to an opportunistic screening setting.

In \cite{pickhardt2013opportunistic}, Pickhardt et al. were able to discriminate manually between patients with osteoporosis and with normal BMD using opportunistic abdomen CTs. 
Recently, several related deep learning based approaches for semi-automated osteoporosis diagnosis have been presented. Ho et al. \cite{ho2021application} and Hsieh et al. \cite{hsieh2021automated} used a deep learning architecture to predict DXA based aBMD from hip X-ray.
Other works like \cite{jang2021prediction} or \cite{yamamoto2020deep} 
detect osteoporosis
directly from image features of X-ray using end-to-end classification networks.
They achieve high classification performance (AUC $> 0.9$) which could even be slightly improved \cite{yamamoto2020deep} by incorporating clinical risk factors (AUC $>0.92$).
However, this diagnosis task it not comparable to the prognosis task that we target in our work.
For prognosis, Hsieh et al. \cite{hsieh2021automated} used their predicted aBMD as input to \frax to predict a 10-year fracture risk.
However, since the performance of this combination is limited by the performance of DXA-based aBMD, they were unable to achieve any improvement over baseline \frax + (DXA-based) aBMD.

Recently, Damm et al. proposed a fully automatic deep learning method to predict hip fractures in women using X-rays from the Study of Osteoporotic Fractures (SOF\footnote{The Study of Osteoporotic Fractures (SOF:) \url{https://sofonline.ucsf.edu}})\cite{Damm2021Sofia}.
They showed that deep learning based methods are able to improve the prognostic performance of classical aBMD based models while maintaining a high degree of automation.
However, they did not investigate additional risk factors and other image modalities as input such as CT.

\section{Method}
\label{sec:method}

We propose an automatic image processing pipeline for the prediction of \textbf{F}uture \textbf{O}steoporotic Fractures \textbf{R}isk in \textbf{M}en for Hips (FORM).
The pipeline consists of a preprocessing, a feature extraction and a risk estimation stage for each patient $x$.
An overview is given in \autoref{fig:pipeline-0}.

\subsection{Preprocessing}
\label{subsec:preprocess}

The proposed method should be able to process 2D X-rays as well as 3D CT scans.
To share both architecture and hyperparameters for both input modalities, we compute 2D projections from the 3D CT scans.
This way, however, most of the 3D structural information from the CT scans is lost.
To fully exploit the 3D information, a native 3D CNN could have been used, but this would have resulted in a much larger memory footprint and thus higher hardware requirements.
Therefore, in this work, we have focused at first on confirming the usefulness of CT image data for predicting future fractures.
In the following, the CT projections will be referred to simply as CT.

A Hough transform is used to detect the QCT calibration phantom that is present in all scans in order to remove it and the underlying table from the image\footnote{In the MrOS study, the phantoms are used to calibrate HU to BMD.
In this work no BMD calibration is performed for a more realistic opportunistic screening setting}.
Since the phantom is always located beneath the patient, the scans can easily be cropped to exclude the phantom and the table.
The cropped 3D scans are then projected onto the coronal plane and re-scaled by a constant value to achieve pixel value range of  $[0, 1]$.
We also investigated a re-scaling per patient to investigate the influence of the scanner HU calibration; results on these normalized CTs (CTN) are reported in the supplement.\\
CT projections and X-ray images $I(x)$ were split into two halves depicting the right and the left hip, respectively; images of the left hip were vertically flipped.
A key point detection CNN inspired by \cite{Damm2021Sofia} were used to detect 12 key points located around the femur.
The key point detector was trained jointly on 1797 X-ray images and 208 CT projections (104 CT, 104 CTN) with manually annotated key point positions.
The key point CNN classifies the image halves into three classes: \emph{complete} (full proximal femur is visible), \emph{incomplete} (proximal femur not completely visible) and \emph{implant}.
A selection of key points is used to crop the image to the proximal femur region (including the trochanter minor and the femoral head).
These cropped images $G(x)$ are included in the dataset if the predicted class is \emph{complete} with a confidence above $0.01$ (X-ray) or $0.2$ (CT).

The risk factors (RF) are additional information about the patient which might improve the hip fracture risk prediction.
As we are not interested on the impact of a single risk factor, but rather whether the information is helpful in combination with image data, we grouped the RF for better referencing:
\emph{Base}, \emph{Multiple}, \emph{aBMD}, \emph{\frax} and \emph{TBS}\cite{schousboe2018predictors}.
The details are summarized in the supplementary.
The \emph{Base} group contains basic patient information like age and BMI.
The \emph{Multiple} group extends \emph{base} and adds additional information from the case history and health background.
This information might not be present in clinical routine but could be acquired from every patient via a questionaire (non-densitometric).
The other groups consists of other well-known risk factors, also including densitometry.
For densitometry, additional imaging and evaluation is required and thus is not suitable for opportunistic screening.
We included these risk factors as a comparison.
52 patients were excluded from the dataset due to missing data for at least one risk factor.

\subsection{Feature Extraction}

We train a CNN as a Feature Extractor with output $r'(x)$ on the cropped femur images $G(x)$ and extract the predicted Global Average Pooling (GAP) Features of the network. These GAP features $I_G(x) \in \mathbb{R}^{2048}$ are used as input for the next pipeline stage.
For the training of the CNN, a ground truth label $F_t(x)$ is needed indicating whether the patient $x$ will fracture by time horizon $t$ (e.g. 10 years) or not. 
All patients with unknown fracture status, e.g., due to death before time horizon $t$,  were excluded from the dataset.
This renders all predicted risks conditional on the patient survival to time horizon $t$.
This is acceptable for an opportunistic screening because we need to screen all patients regardless of whether or not they would survive to $t$\cite{americanClinicalPractise} but might introduce a bias.

FORM has a ResNet50v2 \cite{resnet} backbone pretrained on ImageNet\cite{imagenet} with three additional layers depicted in \autoref{fig:pipeline-1}.
Data was augmented with random flips, zooms and color changes; classes were weighted.
Input image dimensions were 96x96 (CT) or 224x244 (X-ray) pixels.
Training was performed for 50 epochs, a batch size of 36, learning rate of $10^{-4}$, dropout (0.5)\footnote{Implemented in Tensorflow 2.4, source code will be release on publication, experiments executed on Nvidia RTX 3090, inference $<$ 1 second per image}.
A cross-entropy loss was used and input samples were weighted based on their class distribution.
GAP features $G(x)$ were extracted after early stopping (see \autoref{subsec:datasets} for details about the metrics).

\subsection{Risk Estimation}

For the risk estimation $r(x)$, we train a multi layer perceptron (MLP) to predict hip fractures up to the horizon $t$ with the target $F_t(x)$.
Its input can be varied:  GAP-Features $I_G(x)$, RF $I_R(x)$ or both.
Categorical data is one hot encoded and concatenated with normalized continuous data into the input vector $I_R(x)$.
Using many RF together with the high-dimensional GAP features might make the model more prone to overfitting because individual datapoints become more distinct.
This is counteracted by a high dropout rate.
To prevent imbalancing between $I_G(x)$ and $I_R(x)$ due to high dimensional differences (e.g. 2048 vs. 4) an MLP with $128$ and $s \cdot k$ hidden nodes is used to reduce the dimension of the GAP features before concatenation.
Here $k$ is the dimension of $I_R(x)$ and $s$ is a scaling hyperparameter set to $5$. 
Hyperparameters are mostly shared in Feature Extraction and Risk Estimation and the differences are illustrated in  \autoref{fig:pipeline-1}.
An ablation study to inspect the impact of the hyperparameters is included in the supplementary material and discussed in \autoref{subsec:ablation}.

\section{Evaluation}

\subsection{Datasets and Baseline Methods}
\label{subsec:datasets}

\tblStatistics

For training and evaluation, we used the dataset from the Osteoporosis in Men (MrOS) study\footnotetext{\url{https://mrosonline.ucsf.edu}, Update august 2021}.
Patients were followed for more than 10 years.
We used the first hip fracture that occurred after the baseline visit as our primary outcome.
A detailed overview of the datasets statistics for the X-ray and the CT cohort in comparison to the complete study population can be found in \autoref{tbl:statistics}.
The number of included patients $n$ is decreased by about one third of the overall number of patients with available image data due to censoring and excluded image halves (e.g. due to implants).
During a 5 / 10 year follow-up 1.45\% and 3\% of the men suffered a hip fracture, respectively.
This low number of cases limits the generalizability but it is possible to identify trends which repeat across different modalities, horizons and settings.
Therefore, we use the same training validation split based on the patient IDs across all experiments for the respective cohorts.
We used area under the receiver-operator curve (AUC) as the main metric.
A 5-fold cross-validation was used to ascertain the validity of the comparison with the established baselines;
across folds validation means and standard deviations (STD) are reported.
To ensure reproducibilty training was repeated 10 times for deep learning models.
In the ablation studies, we analyzed only one fold across 10 repetitions and report means with their standard errors (SE). 
A two-sided Welch-Test\cite{welch1947} was used to compare the calculated means.

The folds are fixed for all input combinations.
However, since not all patients have both, X-ray and CT data different subgroups of the training and validation data were used for each modality.
This means that the different cohorts are similar but can not be directly compared.
Therefore, we compare across both cohorts with different inputs / models (e.g. \frax).
Matching the X-ray and CT cohorts would have resulted in too few fracture cases.
Sampling was performed on a per-patient basis, ensuring the two image halves for one patient always are in the same fold.

As baselines a Cox Proportional-Hazards Model (Cox) \cite{cox} and \frax was used.
For the Cox model the same input as to our model FORM was used. 
However, the low variance of the high dimensional GAP features lead to a numerical degeneration of the Cox model.
This was circumvented by performing a dimensionality reduction using Principal Component Analysis (PCA) \cite{pca} of the GAP feature space. 
The Cox model was fitted on the training data and used for prediction on the validation data.
Best performing number of PCA components are reported based on the validation set.

\subsection{Results}
\label{subsec:results}

\tblCross
\tblHor

The proposed method (FORM), a Cox Proportional-Hazards Model (Cox) and \frax are compared using a five-fold cross-validation analysis in \autoref{tbl:cross}. 
It can be seen, that the proposed method outperforms Cox and \frax on both cohorts by around 6\%. 
In general, using more risk factors in the GAP feature input improves the prediction; this benefit is slightly higher on the CT cohort. 
Especially for Cox, a high variance without risk factors can be observed which can be credited to one or two folds with significant lower performance (e.g. around 35\% on one fold for the X-ray Cohort).
The number of the PCA components were 2 and 1 for the X-ray and CT cohort, respectively.
While more components lead to improved training scores the values on the validation data degenerate and thus indicate an overfitting.
We checked for the Cox models the variables with a significant impact on the  prediction on at least two folds.
The variables for the X-ray cohort were age, gap features, smoking, fall history and BMI.
For the CT cohort the variables were age, gap features, BMI, cancer and hypertension.

On one fold the power to predict hip fractures were analyzed further by adding comparisons without GAP features and evaluations including densitometric inputs variables in \autoref{tbl:results}.
Across both cohorts, a significant improvement of around 20\% to only using the risk factors group Base or Multiple can be seen.
Moreover, a significant improvement of up to one percent is achieved when adding information about risk factors on X-ray and the vanilla \frax is worse for both cohorts. 
Overall the image-based result are all similar and within a range of around three percent of 80\%.
For the densitometric settings, only the usage of Base risk factors is reported, because further information did not improve the results.
In the comparison of the best non-densitometric model, with densitometric settings FORM still outperform most risk factors or the vanilla \frax+aBMD predictor.
Only X-ray based imaging is up to three percent worse than \frax based predictions.
We see an improvement of using the \frax+Base as input to FORM in comparison to the vanilla \frax predictor.

The ablation study on the CTN for the 10 and 5 year horizon in the supplement shows that changes to the used hyperparameters lead to significant degenerations of up to three percent.
Without input merging the additional risk factors can not be benefically used or even degenerate the results due to overfitting.
A lower dropout rate or a larger model which result in faster overfitting have a similar effect.
On the CTN data in general more risk factors improve the results by up to one percent.

\subsection{Discussion}
\label{subsec:ablation}

We conclude three major results: (1) FORM outperforms Cox and \frax on two image cohorts and  performs similarly or better even if we include densitometric inputs as comparison, (2) only image information can be used for fracture risk prediction but additional risk factors can help the risk estimation and (3) FORM can leverage the combined information of image information and risk factors better than Cox.

Across all experiments FORM outperforms the other non-densitometric models.
Only the densitometric \frax predictor (including aBMD) performs better on the X-ray cohort than FORM.
However, our models do not require additional imaging with DXA. 
Future research could highlight important image regions for the risk estimation or the importance of additional risk factors which could improve the interpretability and therefore the acceptance of the system in clinical routine.

For Cox and FORM in \autoref{tbl:cross}, it can be seen that a fracture risk estimation only based on image information is possible and even outperforms predictions only based on risk factors in \autoref{tbl:results}. 
Using additional risk factors as input can improve the results significantly by up to four percent.
This shows that risk factors are a valid source for additional information but also that a majority of the information is already encoded in a patient's X-ray or CT.

The Cox model performs in the best case similar or worse than \frax but is outperformed by FORM. 
While \frax might use other input variables, the Cox model is trained with the same inputs as FORM. 
We conclude that our model can learn from the high dimensional data better than Cox due to two reason:
first, the Cox model required preprocessed inputs via PCA to prevent degeneration.
Second, the overfitting prevents adding more than one or two PCA components as input.

In Chapter 2.2, we explained that patients were excluded due to early death.
The censored patients cannot be directly evaluated, but their subgroup which survived for at least the first 5 years without a fracture.
The number of false positive predicted patients across 20 repetitions are 3.63\% $\pm$ 0.51\% SE and 5.01\% $\pm$ 0.80\% SE for the validation and censored subset, respectively.
We conclude that our model is performing at least plausible on this subset.

\subsection{Limitations and Future Work}

This study is based solely on the MrOS dataset, which consists only of men and contains an expected low number of incident (future) fractures.
Both issues limit the generalizabilty of our work but can be verified on other datasets in the future. 
The identified trends are supported across different cohorts and settings but have to be confirmed on other studies especially including women and with a higher number of fractured cases.
2D projections were used to share architectures between X-ray and CT, but native 3D architectures should be investigated.
This study can only analyze the benefits of image data for opportunistic screening in a proof-of-concept fashion, since the strict imaging protocols were imposed for the study.
A long term study in  clinical routine is required to evaluate the practicability and the evaluation of questions like how many false positive / negatives are acceptable in clinical routine or how the output probability should be calibrated in regard to the sensitivity / specificity trade-off.
This will be future work.

\section{Conclusion}

We have shown that X-ray and CT data can be automatically analyzed and processed by our method FORM for opportunistic hip fracture prognosis.
We achieved a mean validation AUC of greater than 80\% for 10-year hip fracture risk in a five-fold cross-validation in both cohorts based on radiographic and CT data.
This is signifcantly better than previous methods like Cox or \frax on the same or comparable input.
Even in most cases, with additional densitometric RF, our method is signficantly better.
Overall, we are confident that our method FORM and image input in general are promising candidates for improving the identification of men at high risk of future osteoporotic hip fractures.

\bibliographystyle{splncs04}
\bibliography{lib}

\newpage
\FloatBarrier
\section{Supplement}

\tblRiskFactors

\tblAblation

\tblDatasets

\end{document}